\documentclass[conference]{IEEEtran}
\IEEEoverridecommandlockouts
\usepackage{cite}
\usepackage{amsmath,amssymb,amsfonts}
\usepackage{algorithmic}
\usepackage{graphicx}
\usepackage{textcomp}
\usepackage{xcolor}

\newcommand{\eg}{\emph{e.g.}}
\newcommand{\ie}{\emph{i.e.}}
\newcommand{\etal}{\emph{et~al.}}
\DeclareMathOperator{\E}{\mathbb{E}}
\usepackage{multirow}

\def\BibTeX{{\rm B\kern-.05em{\sc i\kern-.025em b}\kern-.08em
    T\kern-.1667em\lower.7ex\hbox{E}\kern-.125emX}}
\begin{document}

\title{SAR-to-EO Image Translation \\ with Multi-Conditional Adversarial Networks
}

\author{\IEEEauthorblockN{Armando Cabrera}
\IEEEauthorblockA{\textit{DAF MIT AI Accelerator}\\
Cambridge, MA, USA \\
cabreraa@mit.edu}
\and
\IEEEauthorblockN{Miriam Cha}
\IEEEauthorblockA{\textit{MIT Lincoln Laboratory} \\
Lexington, MA, USA \\
miriam.cha@ll.mit.edu}
\and
\IEEEauthorblockN{Prafull Sharma}
\IEEEauthorblockA{\textit{MIT} \\
Cambridge, MA, USA \\
prafull@mit.edu}
\and
\IEEEauthorblockN{Michael Newey}
\IEEEauthorblockA{\textit{MIT Lincoln Laboratory} \\
Lexington, MA, USA \\
michael.newey@ll.mit.edu}
}

\maketitle

\begin{abstract}
This paper explores the use of multi-conditional adversarial networks for SAR-to-EO image translation. Previous methods condition adversarial networks only on the input SAR. We show that incorporating multiple complementary modalities such as Google maps and IR can further improve SAR-to-EO image translation especially on preserving sharp edges of manmade objects. We demonstrate effectiveness of our approach on a diverse set of datasets including SEN12MS, DFC2020, and SpaceNet6. Our experimental results suggest that additional information provided by complementary modalities improves the performance of SAR-to-EO image translation compared to the models trained on paired SAR and EO data only. To best of our knowledge, our approach is the first to leverage multiple modalities for improving SAR-to-EO image translation performance. 
\end{abstract}


\section{Introduction}
In the past ten years, 1.7 billion people have been affected by climate and weather disasters \cite{wdr2020}. According to the World Disaster report published in 2020, climate and weather-related disasters have increased, whereas the recovery time in-between disasters have decreased. Humanitarian Assistance and Disaster Relief (HADR) missions provide time-critical support, such as detecting damaged infrastructures in areas that natural disasters have impacted. Since much of the region affected by disasters is difficult to access, remote sensing offers a powerful tool to enable situational awareness in the area. However, obtaining a continuous time series of images is hindered by weather, clouds, smoke, and other inherent limitations of passive sensors such as electro-optical (EO) sensors that rely on sunlight. Such limitations produce an information gap during or post a disaster event delaying the planning and the course of action needed to save lives.

Synthetic aperture radar (SAR) is an active sensor that can collect high-resolution images with relative invariance to weather and lighting conditions. Due to its ability to produce images in all weather and lighting conditions, SAR imaging has advantages in HADR compared to optical systems. Despite SAR's benefits, visual inspection of SAR images is challenging due to the large dynamic range and low spatial correlation. Contrast sensitivity in the human visual system is poor when the spatial correlation is low and significantly degrades at high-frequency \cite{human_vision}. Furthermore, SAR imagery contains radar-specific geometric distortions such as foreshortening, layover, and shadow, making interpreting SAR imagery non-intuitive. The challenges of interpreting SAR imagery for new analysts or analysts who do not regularly work with SAR result in a delay in sending the latest information to the decision-maker.

It would be desirable to have interpretable SAR images in the optical domain while keeping the all-weather and all-lighting high-resolution imaging capability. Fortunately, this is exactly what the recently proposed conditional generative adversarial network (CGAN) does when applied to translating EO images from SAR images. Conditioning gives a means to control the generative process that the original generative adversarial network (GAN) lacks. Isola \etal \cite{pix2pix2017} introduce \texttt{pix2pix}, a CGAN architecture extended on deep convolutional generative adversarial network to do image-to-image translation on paired training data. With generator and discriminator nets conditioned on the input image (\eg~sketches), output images (\eg~photographs) corresponding to the input image can be generated. Building on the \texttt{pix2pix} framework, Zhu \etal \cite{cyclegan2017}  introduce Cycle-Consistent Adversarial Networks (\texttt{CycleGAN}) that learn the image-to-image mapping without paired input-output training examples.

Several recent papers have used variations of \texttt{pix2pix} and \texttt{CycleGAN} for SAR-to-EO translation. Schmitt \etal \cite{schmitt2018sen12} present preliminary results applying \texttt{pix2pix}  to SAR-to-EO translation. Toriya \etal \cite{toriya}  extend the application of the translation network for multimodal image alignment. Wang \etal \cite{wang}  and Reyes \etal \cite{reyes}  use variants of \texttt{CycleGAN}. Wang \etal \cite{wang} incorporate mean squared error loss in addition to cycle-consistency loss in \texttt{CycleGAN} and show improvement over the vanilla \texttt{pix2pix} and \texttt{CycleGAN}. Using CGANs, these papers have achieved impressive results but have focused on conditioning the CGAN only on the input SAR.

In this paper, we explore SAR-to-EO image translation in the multi-conditional setting. To do this, we condition the CGAN on additional information such as maps from OpenStreetMap, infrared (IR), and latitude and longitude coordinates along with the input SAR. Recently, large-scale multimodal datasets that include additional information to SAR and EO images have been generated. Yet, current state-of-the-art algorithms have focused primarily on learning from SAR images only. We hypothesize that each modality is noisy and incomplete. Therefore, incorporating multiple complementary modalities can help in learning the better mapping between SAR and EO images. To the best of our knowledge, this work represents the first attempt to leverage additional complementary modalities (\eg~maps from OpenStreetMap and lat/lon coordinates) for SAR-to-EO translation. We evaluate the proposed multi-conditional GAN on various multimodal remote sensing datasets and share points of success and modes of failure. Our qualitative and quantitative results on SpaceNet6, DFC2020, and SEN12MS suggest that adopting additional conditioning helps improve visual realism in SAR-to-EO image translation, especially on better transferring boundaries of edges.

The rest of this paper is organized as follows. Section~\ref{sec:bg} provides background on the generative adversarial net, covering its variants and mathematical formulations. In Section~\ref{sec:method}, we describe the proposed multi-conditional GAN. Section~\ref{sec:exp} presents an experimental evaluation of the propose approach on the SpaceNet6, DFC2020, and SEN12MS datasets. Section~\ref{sec:conc} concludes the paper.

\section{Background}
\label{sec:bg}

\subsection{Generative adversarial nets (GAN)}
Generative adversarial nets (GAN) \cite{goodfellow2014} are composed of two models, a generative model $G$ and a discriminative model $D$ that are simultaneously trained in a two-player mini-max game with the following objective function:
\begin{equation} \label{eq:gan} \small
\min_G\max_D \E_{\mathbf{y}}[\log D(\mathbf{y})] + \E_{\mathbf{z}}[\log(1-D(G(\mathbf{z})))] 
\end{equation}
where $\mathbf{y}$ is a real image from the real data distribution $p_{\text{y}}$ and $\mathbf{z}$ is a noise vector (\eg, Gaussian distribution) with a prior distribution $p_{\mathbf{z}}$. Here, the objective of $G$ is to produce a data estimate $\hat{\mathbf{y}}$ to the real $\mathbf{y}$ using the noise vector $\mathbf{z}$. Meanwhile, $D$ optimizes to distinguish the real image $\mathbf{y}$ and a generated image $\hat{\mathbf{y}}=G(\mathbf{z})$.  In practice, $G$ and $D$ are typically implemented as neural nets. We can train $G$ and $D$ by backpropagation.


\subsection{Conditional generative adversarial nets (CGAN)}
Conditional generative adversarial nets (CGAN) \cite{mirza2014conditional} extend GAN by conditioning $G$ and $D$ on side information. For CGAN, Eq. (\ref{eq:gan}) is rewritten as 
\begin{equation} \label{eq:cgan} \small
\min_G\max_D \E_{\mathbf{x},\mathbf{y}}[\log D(\mathbf{x},\mathbf{y})] + \E_{\mathbf{x},\mathbf{z}}[\log(1-D(\mathbf{x},G(\mathbf{x},\mathbf{z})))]
\end{equation} 
where the extra information $\mathbf{x}$ can be a class label. In neural net implementation, $G$ takes in $(\mathbf{x},\mathbf{z})$ as a joint representation that concatenates $\mathbf{z}$ and $\mathbf{x}$ into a single input vector. Similarly, another joint representation is used to train $D$. 

\texttt{pix2pix} uses a variant of CGAN and provides the noise vector in the form of dropout, not as an input to the generator. Therefore, the objective function for \texttt{pix2pix} becomes:
\begin{equation} \label{eq:pix2pix} \small
\min_G\max_D \E_{\mathbf{x},\mathbf{y}}[\log D(\mathbf{x},\mathbf{y})] + \E_{\mathbf{x}}[\log(1-D(\mathbf{x},G(\mathbf{x})))]
\end{equation} 
Here, the joint representation is taken as a channel-wise concatenation.



\section{Method}
\label{sec:method}

\begin{figure}[t]
\centering
\includegraphics[width=.5\textwidth]{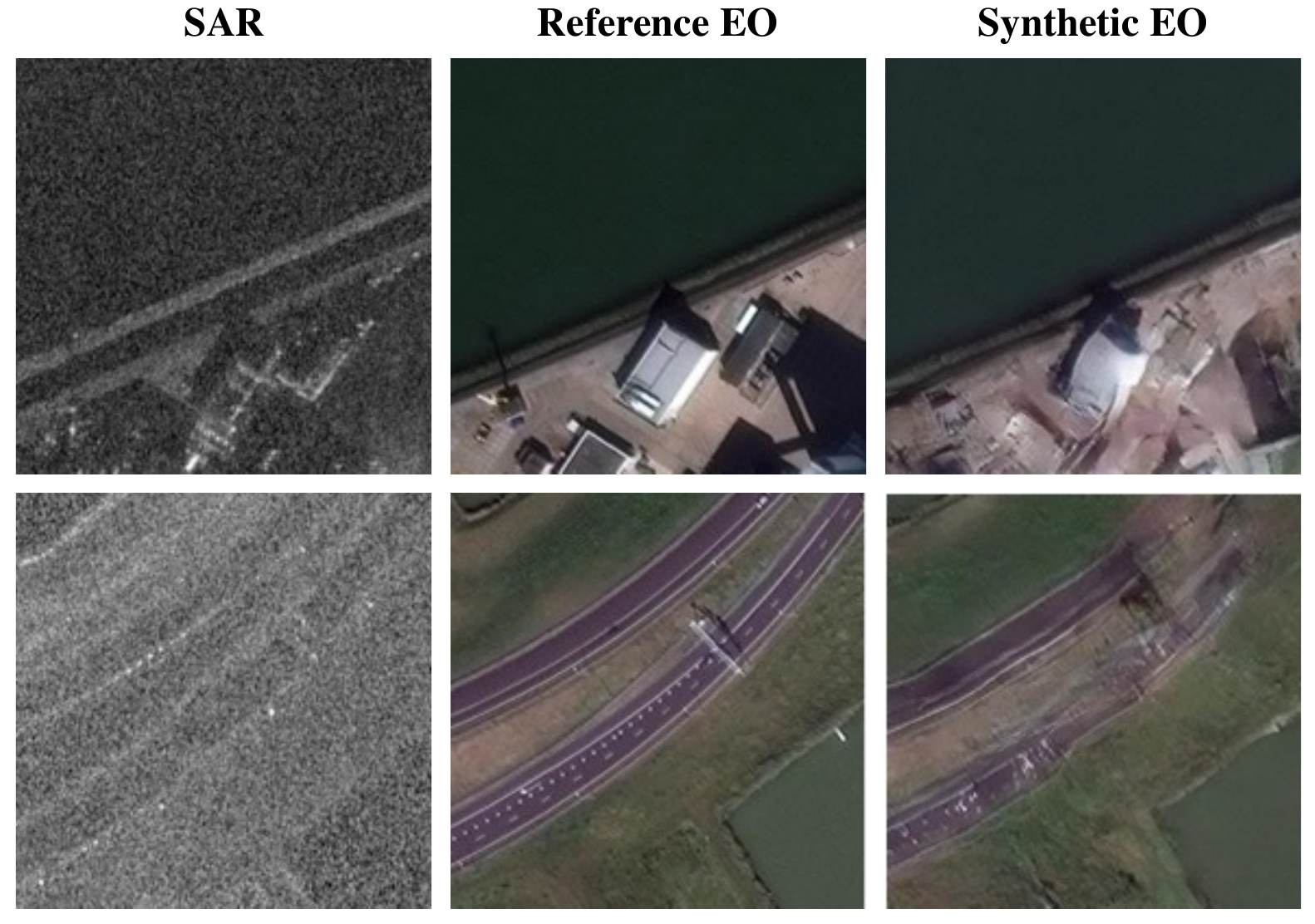}
\caption{Details in boundaries of madmade objects (\eg~roads, buildings) become blurry due to an ill-posed nature of the SAR-to-EO image translation. This motivates us to provide additional information to the CGAN architecture.}
\label{fig:motivation}
\end{figure}

We aim to further improve SAR-to-EO image translation by providing additional information to the CGAN architecture. Due to a one-to-many mapping in SAR-to-EO image translation, a single SAR image can correspond to multiple possible EO images. For such reason, SAR-to-EO image translation by CGAN is meant to generate a plausible looking EO image, which corresponds to the input SAR instead of generating a ground truth. However, as the ambiguity of mapping between SAR and EO images is high, conditioning the CGAN only on a single input SAR image often results in poor results. This is illustrated in Figure~\ref{fig:motivation} where the ill-posed nature of the SAR-to-EO image translation is particularly pronounced for sharp edges, for which boundaries of manmade objects become blurry. 

Recently, large-scale multimodal datasets that include additional information to SAR and EO images have been generated \cite{sen12ms,dfc2020,spacenet}. We explore SAR-to-EO image translation in multi-conditional setting to learn better mapping between SAR and EO images. We propose a simple improvement in CGAN by conditioning the model on an additional information as
\begin{equation} \label{eq:cgan} \small
\min_G\max_D \E_{\mathbf{x,s,y}}[\log D(\mathbf{x},\mathbf{s},\mathbf{y})] + \E_{\mathbf{x,s}}[\log(1-D(\mathbf{x},\mathbf{s},G(\mathbf{x},\mathbf{s})))]
\end{equation}
where $\mathbf{s}$ is an additional complementary modality to the input SAR such as MODIS land cover map and OpenStreetMap. For our proposed multi-conditional GAN, we adopt the \texttt{pix2pixHD} framework \cite{pix2pixhd}, which improves \texttt{pix2pix} by using a coarse-to-fine generator, a multi-scale discriminator architecture, and a discriminator feature matching loss. As our images are of size $256 \times 256$, we only use the global generator in \texttt{pix2pixHD}.


\section{Experiments}
\label{sec:exp}

\subsection{Dataset}
We evaluate the proposed method on multimodal datasets including SEN12MS \cite{sen12ms}, DFC2020 \cite{dfc2020}, and SpaceNet6 \cite{spacenet}. The SEN12MS dataset includes dual-polarization C-band SAR collected by Sentinel-1 and 13-band multispectral images by Sentinel-2. SEN12MS comprises 180,662 co-located SAR and multispectral images with corresponding MODIS land cover maps. The DFC2020 dataset provides additional 12,289 Sentinel-1 and Sentinel-2 image pairs at 10m resolution. EO images are processed by taking the RGB bands from Sentinel-2. SpaceNet6 has 4,301 quad-pol X-band SAR and EO image pairs with $900 \times 900$ image size at 0.5m resolution.

\begin{figure}[t]
\centering
\includegraphics[width=.5\textwidth]{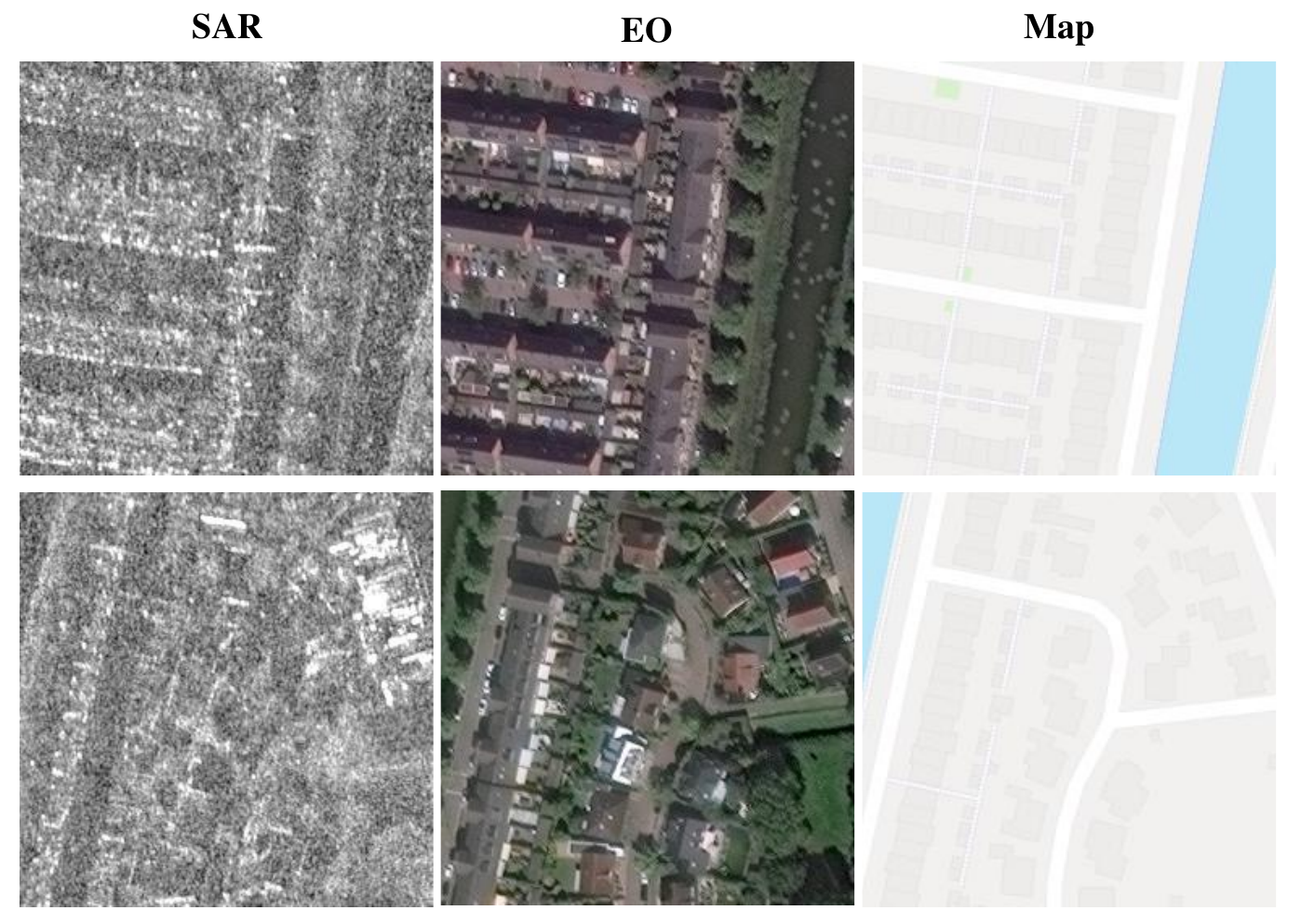}
\caption{Example images of SAR, EO, and corresponding maps scraped from OpenStreetMap}
\label{fig:osm}
\end{figure}

\subsection{Experimentation Setups Details}
For dual-pol SAR, we compose a three-channel SAR image (R: $VV$, G: $VH$, B: $|VV|/|VH|$). All four polarizations are taken for quad-pol SAR. For visual purposes, we plot single-channel SAR in all presented figures. For SEN12MS and DFC2020, we apply the following steps to remove noisy training data to ensure cloud-free training images. First, we convert the images from RGB to HSV. Then, we only keep the images with the mean value of its V channel greater than 0.2.  For the SpaceNet 6 dataset, we chip out the images to $256 \times 256$ and remove the occluded or masked chips by more than 10\%. After the process, the number of images becomes 17,122. Additionally, we scrape corresponding maps from OpenStreetMap (OSM) as samples shown in Figure~\ref{fig:osm}. We split training and testing data by 80/20.


%
%


All models are trained with mini-batch stochastic gradient descent with a mini-batch size of 4 and 400 epochs. We adopt the ADAM optimizer with a learning rate 0.0002 and momentum parameters $\beta_1 = 0.5$ and $\beta_2 = 0.999$. Following Wang \etal~\cite{pix2pixhd}, we use 3 discriminators and feature matching loss parameter $\lambda=10$.

\subsection{Qualitative Evaluation}

\begin{figure}[t]
\centering
\includegraphics[width=.5\textwidth]{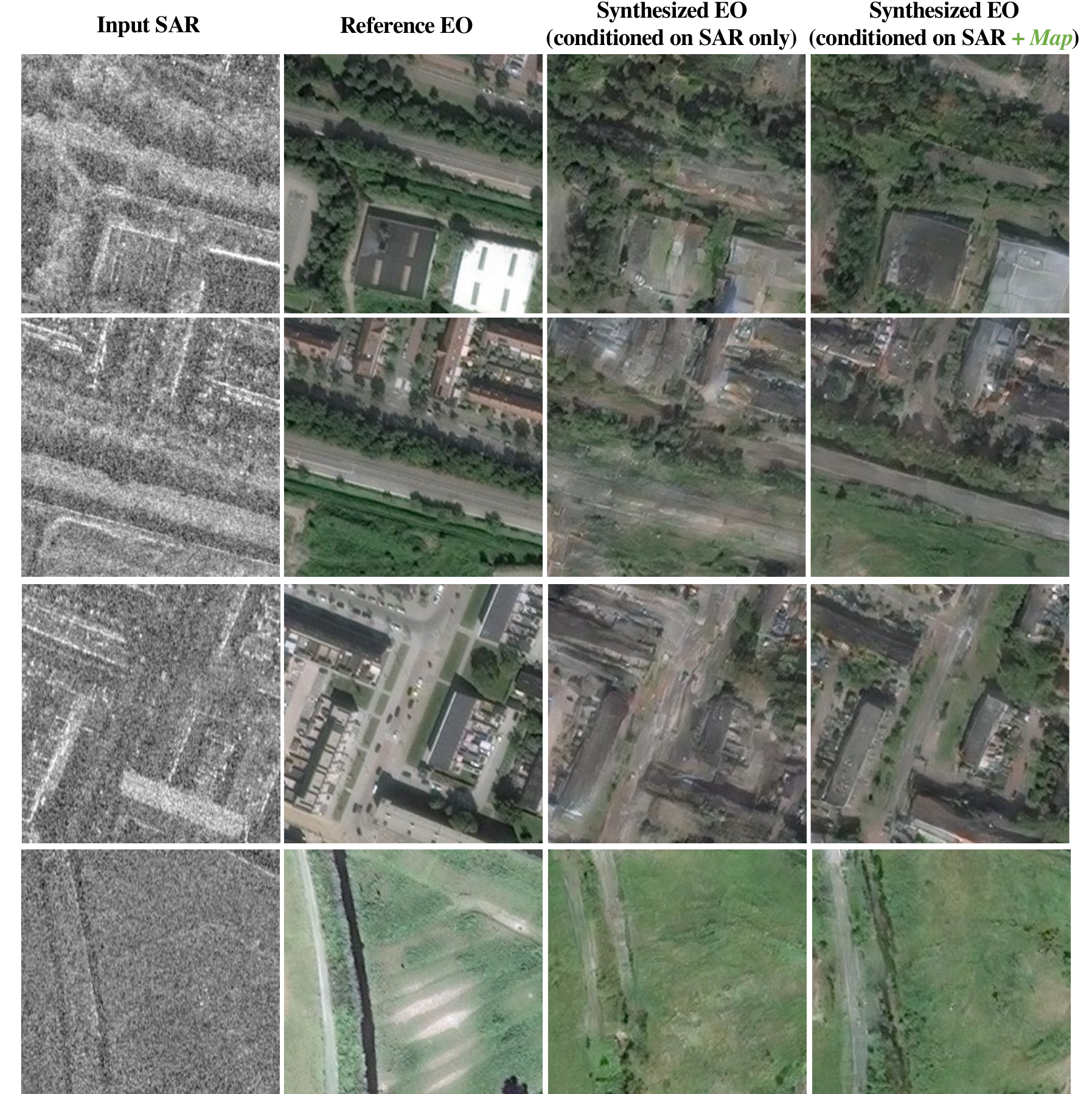}
\caption{SpaceNet6 results. Input SAR (left), Reference EO (middle left), Synthesized EO conditioned on input SAR only (middle right), Synthesized EO conditioned on SAR and corresponding map}
\label{fig:spacenet_result}
\end{figure}

\begin{figure}[t]
\centering
\includegraphics[width=.5\textwidth]{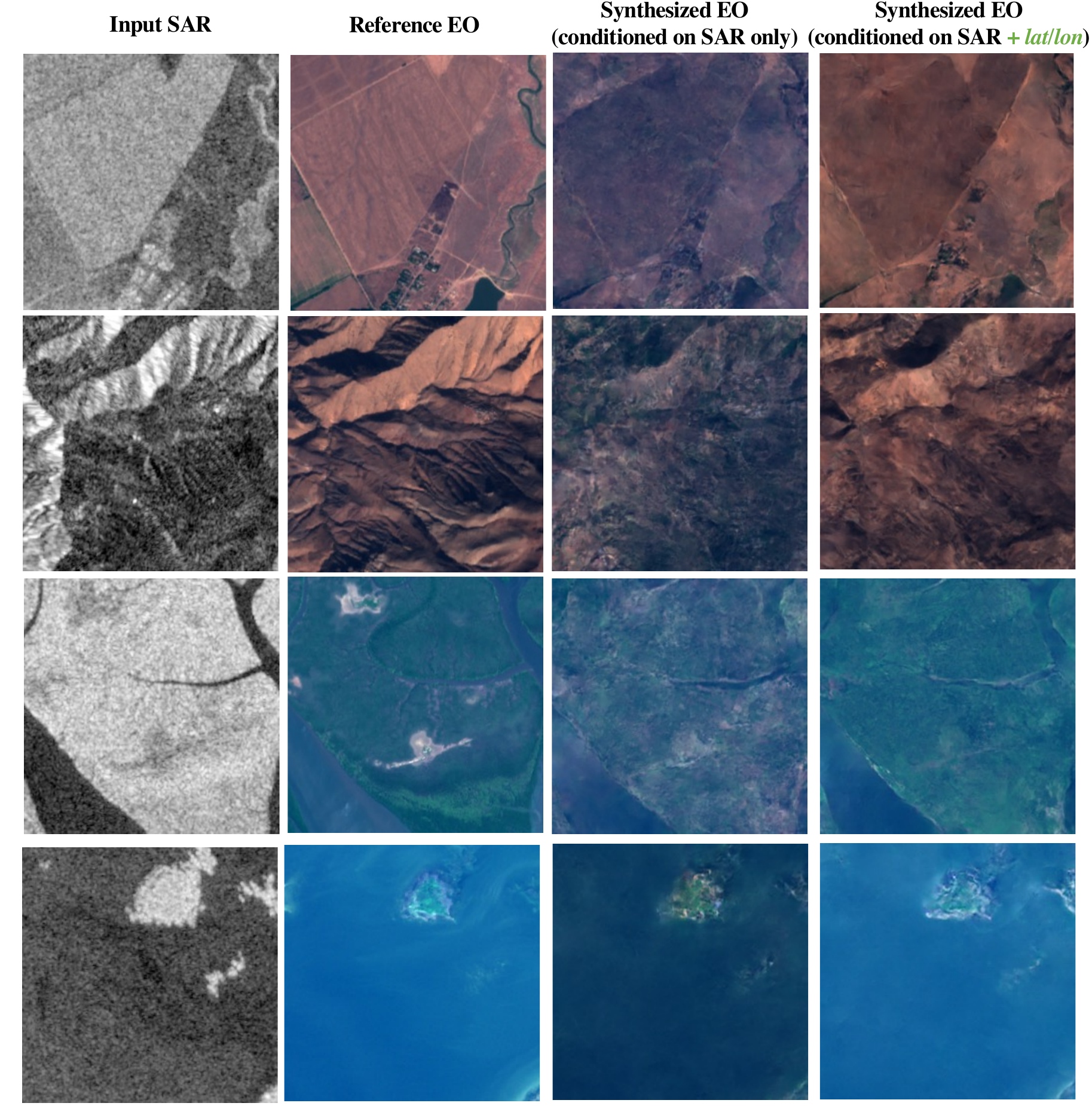}
\caption{SEN12MS results. Input SAR (left), Reference EO (middle left), Synthesized EO conditioned on input SAR only (middle right), Synthesized EO conditioned on SAR and corresponding center latitude and longitude coordinates}
\label{fig:sen12ms_result}
\end{figure}

\begin{figure}[t]
\centering
\includegraphics[width=.5\textwidth]{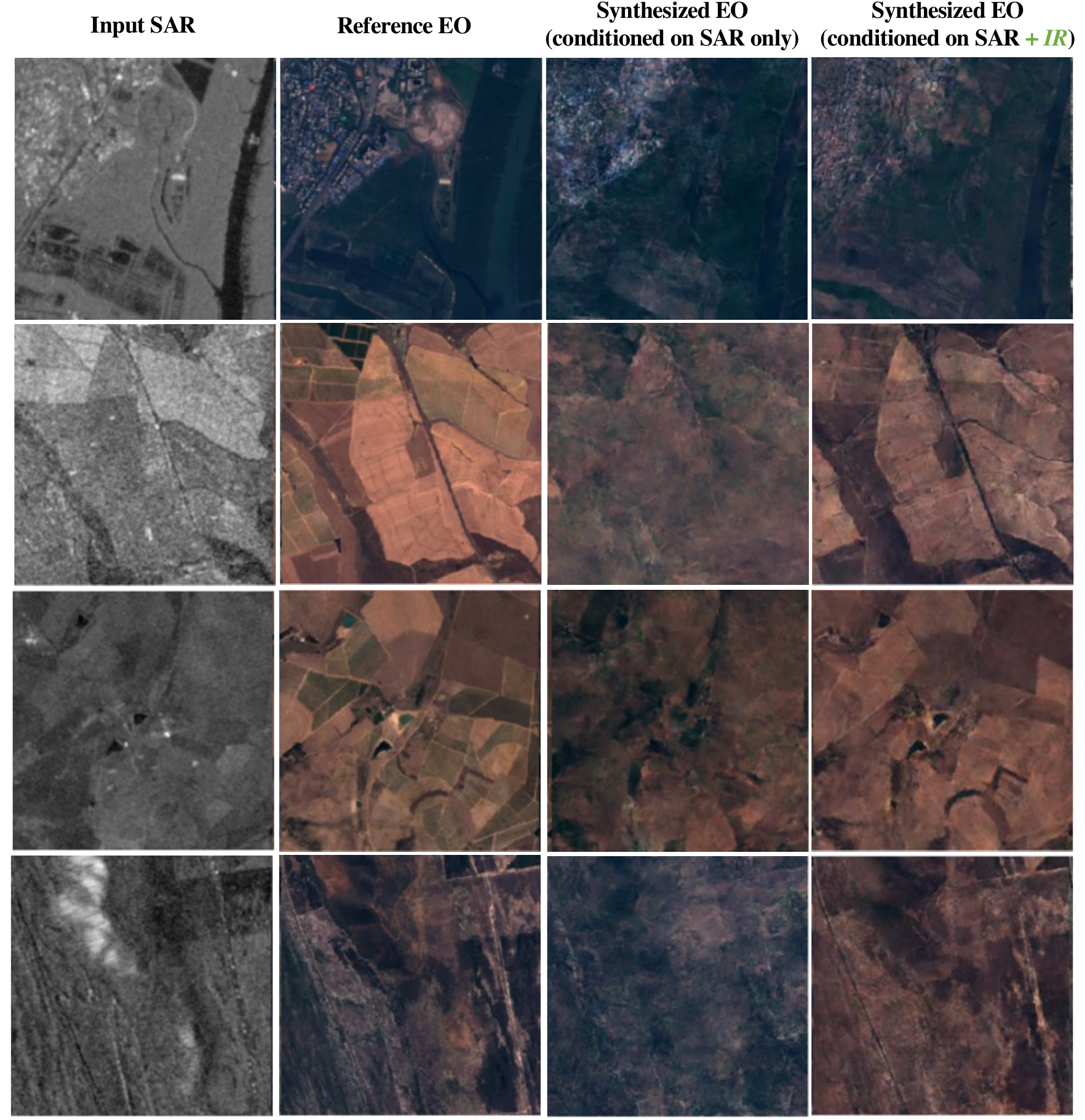}
\caption{DFC2020 results. Input SAR (left), Reference EO (middle left), Synthesized EO conditioned on input SAR only (middle right), Synthesized EO conditioned on SAR and corresponding IR}
\label{fig:dfc_result}
\end{figure}

Figure~\ref{fig:spacenet_result} shows qualitative results on the SpaceNet 6 dataset comparing between the baseline that was conditioned on SAR only and the proposed multi-conditional GAN conditioned on SAR and OpenStreetMap. In these results, we notice that the synthesized EO images with the multi-conditional GAN are visually more convincing than the baseline images. From the first row example, with the multi-conditional GAN, we observe that buildings on the lower right side of the image have sharper linear features and better-represented details. From the second row example, roads have a better linear structure than the baseline. For the final row, looking at the SAR imagery, we see two parallel lines that look similar to each other. However, we observe from the EO that one is a road and the other is a water feature. The baseline result cannot distinguish between the two features, whereas our approach attempts to generate two different features.

Both SEN12MS and DFC2020 capture mostly rural scenes. Figure~\ref{fig:sen12ms_result} presents SEN12MS results utilizing the image's center latitude and longitude coordinates as an additional modality. We spatially replicate the coordinates and concatenate them with the input SAR image. For DFC2020 results, as shown in Figure~\ref{fig:dfc_result}, an infrared (IR) modality is used to condition along with the input SAR. The SEN12MS and the DFC results show that multi-conditioning transfers color and texture better than the baseline result.

\subsection{Quantitative Evaluation}

\begin{table}[]
\centering
\caption{Comparison of baseline (conditioned on SAR only) and multi-conditional GAN (conditioned on SAR and additional modality). Multi-conditional GAN for SpaceNet6 conditions on SAR and map, DFC2020 conditions on SAR and IR, and SEN12MS conditions on SAR and lat/lon.}
\label{tab:result_table}
\begin{tabular}{ccccc}
\multicolumn{2}{c}{}                           & PSNR           & SSIM          & LPIPS         \\ \hline \hline
\multirow{2}{*}{SpaceNet6} & Baseline          & 15.99          & 0.17          & 0.45          \\ \cline{2-5}
                       & Multi-conditional & \textbf{16.20} & \textbf{0.18} & \textbf{0.43} \\ \hline \hline
\multirow{2}{*}{DFC2020}   & Baseline          & 28.90          & 0.70          & 0.12          \\ \cline{2-5}
                       & Multi-conditional & \textbf{29.90} & \textbf{0.79} & \textbf{0.09} \\ \hline \hline
\multirow{2}{*}{SEN12MS}   & Baseline          & 16.64          & 0.30          & 0.48          \\ \cline{2-5}
                       & Multi-conditional & \textbf{17.00} & \textbf{0.32} & \textbf{0.45} \\ \hline
\end{tabular}
\end{table}

In Table~\ref{tab:result_table}, we present quantitative results comparing the baseline and the proposed multi-conditional GAN on three metrics: Peak signal-to-noise ratio (PSNR), Structural Similarity Index (SSIM) \cite{ssim}, and Learned Perceptual Image Patch Similarity (LPIPS) \cite{lpips}. SSIM quantifies the similarity between two images based on luminance, contrast, and structure. SSIM ranges between 0 and 1, where 1 is the highest performing. LPIPS relates to how humans perceive image similarity by using features derived from deep networks. LPIPS ranges between 0 and 1, where 0 is the highest performing.

For SpaceNet6, we use map images scraped from OpenStreetMap as an additional modality, IR modality for DFC2020, and latitude and longitude for the SEN12MS dataset. We observe the multi-conditional method performs better than the baseline on all datasets. With the multi-conditioning approach, LPIPS improve by 0.02 for the Spacenet6 test case, DFC2020 by 0.03, and SEN12MS by 0.03. We have also tested multi-conditioning using a MODIS land cover map provided by the SEN12MS dataset. However, the result came out poor. As discussed in the next subsection, the land cover map is low-resolution and hardly reflects EO or SAR attributes.

\subsection{Modes of Failure}
\begin{figure}[t]
\centering
\includegraphics[width=.5\textwidth]{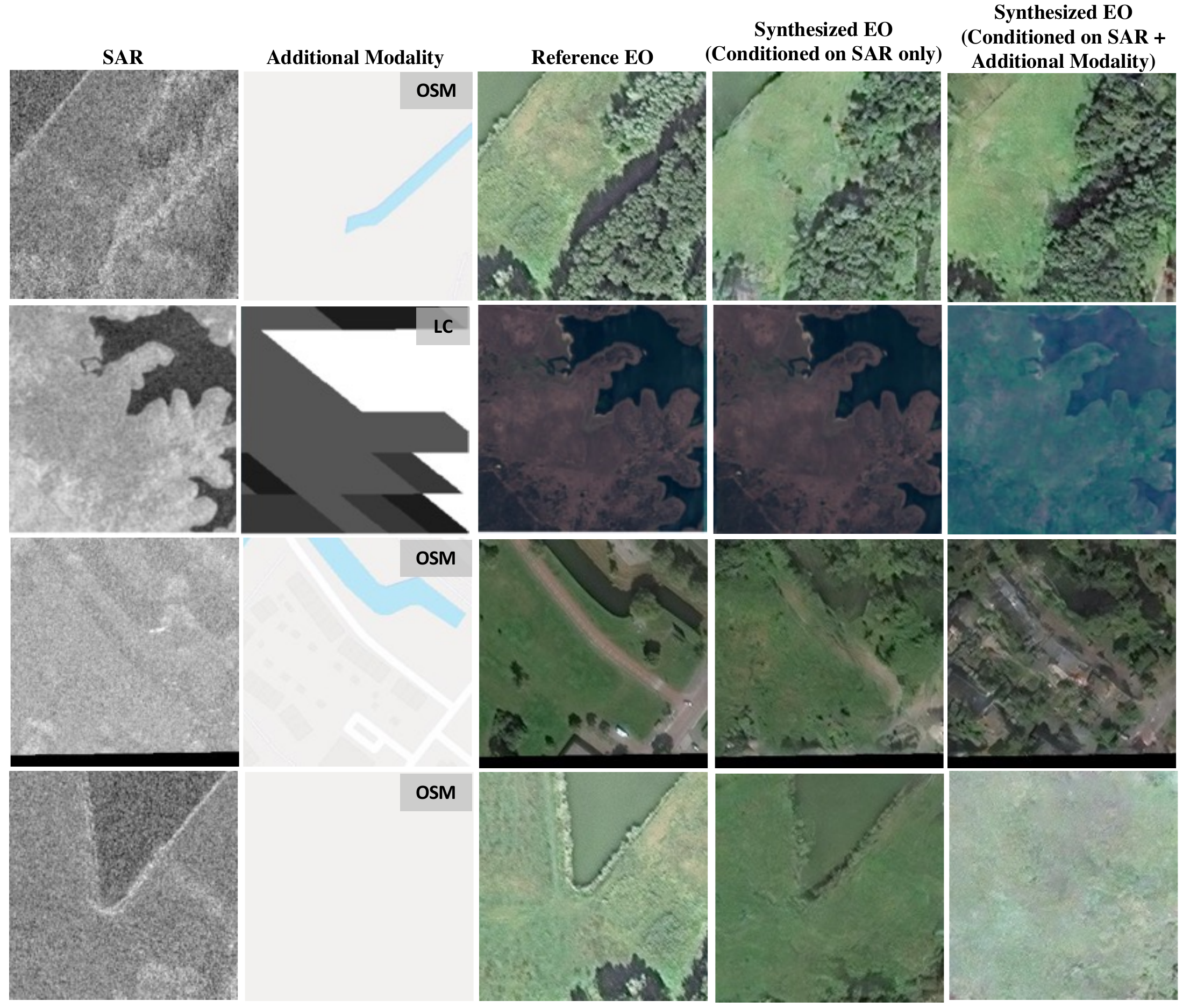}
\caption{Example images from SpaceNet6 and SEN12MS illustrating modes of failure in multi-conditional GAN. Input SAR (left), additional modality (middle left), reference EO (middle), synthesized EO conditioned on input SAR only (middle right), synthesized EO conditioned on SAR and corresponding complimentary modality (right). For all four examples, the baseline performs better than the multi-conditional GAN.}
\label{fig:failure_modes}
\end{figure}

The model assumes an aligned dataset when training an image-to-image translation model based on CGAN (\ie~\texttt{pix2pix}, \texttt{pix2pixHD}). However, a perfectly co-collected multimodal dataset is rare to come by for remote sensing due to physical limitations. Although most images are relatively well aligned, we find that deviation from the assumption due to noise in alignment can cause failure examples in SAR-to-EO image translation. In Figure~\ref{fig:failure_modes}, we show several examples of misaligned images negatively affecting the multi-conditional GAN performance.

The first row is when the additional modality does not match SAR or EO. Both SAR and EO have a triangular water feature on the top left corner of the image, whereas the map image from OSM does not. For the second row, the additional modality is a MODIS land cover map provided by the SEN12MS dataset. We find that the land cover maps are coarse; therefore, conditioning on the low-resolution map hurts the translation performance. In the third-row example, the OSM contains buildings on the left side of the image. We can hypothesize that SAR and EO were taken before the construction of the new facilities. Due to the temporal difference between the SAR and the map, the multi-conditioning model generates buildings not present in the input SAR. Similarly, the mismatch between the SAR and the map negatively affects the last example's multi-conditional GAN performance. Designing multi-conditional GAN that can handle aligned as well as noisy, misaligned training data is an important question left open by the present work.

\section{Conclusion}
\label{sec:conc}

This paper explores the use of multi-conditional adversarial networks for SAR-to-EO image translation. Previous methods condition adversarial networks only on the SAR as input. We show that incorporating multiple complementary modalities such as OpenStreetMap and IR modality can improve SAR-to-EO image translation, especially preserving sharp edges of manufactured objects. We demonstrate the effectiveness of our approach on a diverse set of datasets, including SEN12MS, DFC2020, and SpaceNet6. Our experimental results suggest that additional information provided by complementary modalities improves the performance of SAR-to-EO image translation compared to the models trained on paired SAR and EO data only.  Our approach is the first to leverage multiple modalities for improving SAR-to-EO image translation performance. Our generated images are not meant to be ground truth but a plausible view of the ground truth as GANs hallucinate what they generate. It is suggested to use the generated images as an additional resource when analyzing SAR imagery. In the future, we plan to evaluate our generated images using downstream performance tasks such as building detection.

\section*{Acknowledgment}
The authors wish to acknowledge the following individuals for their contributions and support: Erick Pound, Dr. David Burke, Randy Gordon, Tucker Hamilton, Dr. Phillip Isola,  Michael Kanaan,  Dr. Frank W. King, Victor Lopez,  Kyle Palko, Dr. Taylor Perron, John Radovan, Christopher Szul, Steven Zech. Research was sponsored by the United States Air Force Research Laboratory and the United States Air Force Artificial Intelligence Accelerator and was accomplished under Cooperative Agreement Number FA8750-19-2-1000. The views and conclusions contained in this document are those of the authors and should not be interpreted as representing the official policies, either expressed or implied, of the United States Air Force or the U.S. Government. The U.S. Government is authorized to reproduce and distribute reprints for Government purposes notwithstanding any copyright notation herein. 


\bibliographystyle{IEEEtran}
\bibliography{paper}

\end{document}